\documentclass[manuscript,screen]{acmart}
\usepackage{multirow}
\usepackage{graphicx}

\usepackage{xspace}

\makeatletter
\DeclareRobustCommand\onedot{\futurelet\@let@token\@onedot}
\def\@onedot{\ifx\@let@token.\else.\null\fi\xspace}

\def\eg{e.g\onedot} 

\def\ie{i.e\onedot}

\AtBeginDocument{%
  }

\setcopyright{acmlicensed}
\copyrightyear{2025}
\acmYear{2025}
\acmDOI{XXXXXXX.XXXXXXX}

\acmVolume{37}
\acmNumber{4}
\acmArticle{111}
\acmMonth{8}

\acmISBN{978-1-4503-XXXX-X/2018/06}

\begin{document}

\title{SQUARE: Semantic Query‑Augmented Fusion and Efficient Batch Reranking for Training-free Zero-Shot Composed Image Retrieval}

\author{Ren-Di Wu}
\authornote{Both authors contributed equally to this research.}
\email{m314706017.mg14@nycu.edu.tw}
\orcid{0009-0007-4564-8790}
\author{Yu-Yen Lin}
\authornotemark[1]
\email{m124020039@student.nsysu.edu.tw}
\orcid{0009-0008-1515-0366}
\author{Huei-Fang Yang}
\authornote{Corresponding author.}
\orcid{0000-0001-8261-6965}
\email{hfyang@mail.cse.nsysu.edu.tw}
\affiliation{%
  \institution{National Sun Yat-sen University}
  \city{Kaohsiung 804201}
  \country{Taiwan}
}


\renewcommand{\shortauthors}{Wu et al.}


\begin{abstract}
Composed Image Retrieval (CIR) aims to retrieve target images that preserve the visual content of a reference image while incorporating user-specified textual modifications.
Training-free zero-shot CIR (ZS-CIR) approaches, which require no task-specific training or labeled data, are highly desirable, yet accurately capturing user intent remains challenging.
In this paper, we present SQUARE, a novel two-stage training-free framework that leverages Multimodal Large Language Models (MLLMs) to enhance ZS-CIR.
In the Semantic Query-Augmented Fusion (SQAF) stage, we enrich the query embedding derived from a vision-language model (VLM) such as CLIP with MLLM-generated captions of the target image. These captions provide high-level semantic guidance, enabling the query to better capture the user’s intent and improve global retrieval quality. 
In the Efficient Batch Reranking (EBR) stage, top-ranked candidates are presented as an image grid with visual marks to the MLLM, which performs joint visual-semantic reasoning across all candidates.
Our reranking strategy operates in a single pass and yields more accurate rankings. 
Experiments show that SQUARE, with its simplicity and effectiveness, delivers strong performance on four standard CIR benchmarks.
Notably, it maintains high performance even with lightweight pre-trained , demonstrating its potential applicability.


\end{abstract}




\maketitle


\section{Introduction}
In Composed Image Retrieval (CIR)~\cite{chen-cvpr20-VAL,du-tomm25-survey,lee-cvpr21-CoSMo,vo-cvpr19-TIRG}, users express their intent through a multimodal query that pairs a reference image with a user-provided text describing the desired modifications, aiming to retrieve a target image that preserves the core visual content of the reference while incorporating the specified changes.
This multimodal formulation enables users to articulate their search intent more precisely and flexibly than unimodal queries, such as pure image-based search~\cite{su-tcsvt21-AHBN} or text-to-image retrieval~\cite{cao-ijcai22-suervey,qu-sigir21-DIME,rao-sigir22-ITR}.
While more expressive, the multimodal query also introduces a significant challenge for CIR: accurately inferring the user's compositional intent from both the visual and textual cues. 
Traditionally, this is addressed via supervised learning on annotated triplets of (\textit{a reference image}, \textit{a text modifier}, \textit{a target image})~\cite{vo-cvpr19-TIRG, chen-cvpr20-VAL, lee-cvpr21-CoSMo, baldrati-tomm24-CLIP4Cir}. However, since constructing such datasets is labor-intensive and difficult to scale, zero-shot CIR (ZS-CIR)~\cite{baldrati-iccv23-SEARLE, gu-cvpr24-LinCIR, karthik-iclr24-CIReVL, saito-cvpr23-Pic2Word, tang-aaai24-Context-I2W} has emerged as a recent research focus, aiming to perform retrieval without task-specific training. Among the various paradigms for ZS-CIR, we, in particular, study the training-free solutions. 

\begin{figure}[tb]
    \centering
    \includegraphics[width=1\textwidth]{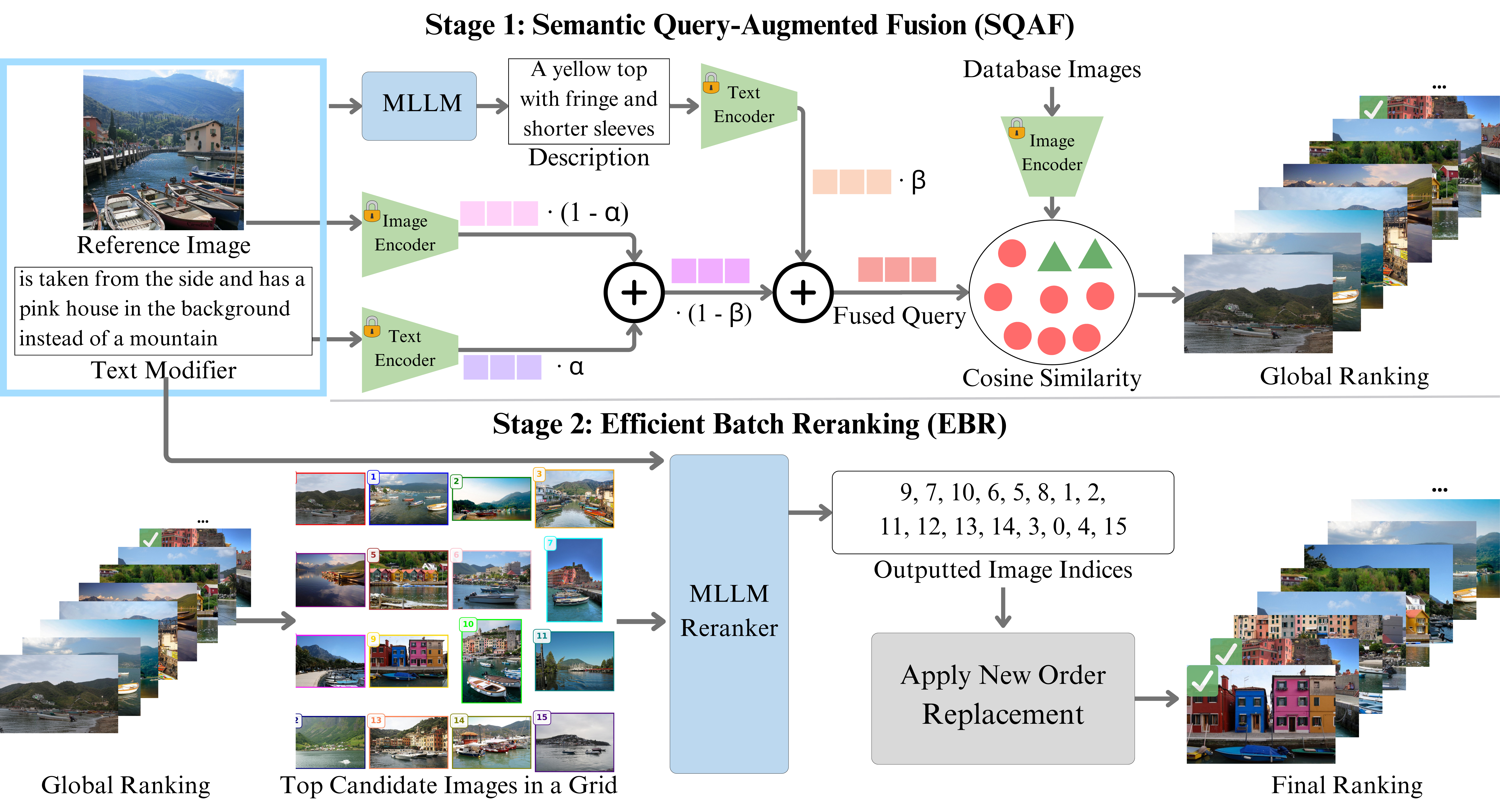}
    \caption{Overview of SQUARE, our training-free framework for zero-shot composed image retrieval (ZS-CIR). SQUARE takes a coarse-to-fine strategy. In stage 1, Semantic Query-Augmented Fusion (SQAF) enriches the embedding-based composed query, formed by fusing the embeddings of the reference image and modification text from a VLM, by incorporating an MLLM-generated target image caption. This addition improves retrieval accuracy while also making the retrieval process interpretable.
    In stage 2, Efficient Batch Reranking (EBR) leverages the MLLM’s multimodal reasoning to refine the ranking. The top-$K$ candidate images from SQAF's output are arranged in a grid, with each image annotated with a distinct label and a bounding box. This presentation allows the MLLM to assess all candidates jointly in a single forward pass, yielding faster inference and more precise reranking.}
    \Description{A two-stage composed image retrieval framework shown from top to bottom. On the top, the MLLM generates a target caption from a reference image and text modifier, which is fused with CLIP embeddings for initial retrieval (SQAF). On the bottom, the top-k retrieved candidates are organized into a grid and reranked by the MLLM through visual-text reasoning (EBR).}
    \label{fig:method}
\end{figure}

To achieve ZS-CIR in a training-free manner, recent studies have leveraged the visual–textual alignment learned by large pretrained models. Vision–language models (VLMs), such as CLIP and BLIP, embed images and text into a shared semantic space, enabling the direct combination of visual and textual cues without additional training. A common strategy is to construct the composed query in this space by manipulating the embeddings of the reference image and the modification text, for example, via a weighted combination~\cite{wu-taai24-WeiMoCIR} or spherical linear interpolation~\cite{jang-eccv24-Slerp}.  
While such embedding-based approaches are simple and efficient, they often treat the modification text as a coarse signal, underutilizing its rich semantic structure. \textit{This raises the question: Can we better capture and integrate the semantic nuances to improve retrieval performance?}

A complementary line of work reframes ZS-CIR as a text-to-image retrieval task~\cite{karthik-iclr24-CIReVL,park-smc25-MCoT-RE,tang-cvpr25-OSrCIR,yang-sigir24-LDRE} leveraging the multimodal reasoning capabilities of multimodal large language models (MLLMs) or the natural language understanding strengths of LLMs. 
The core idea is to utilize these models to generate a detailed target image caption that explicitly encodes the intended compositional reasoning, and then retrieve images using this caption within a VLM’s semantic space.
Although such captions often capture user intent well, retrieval accuracy can be limited by the quality, specificity, and faithfulness of the generated descriptions. 
Moreover, with the rapid progress in MLLMs, \textit{beyond caption generation, can these models be further exploited to directly enhance retrieval?}

In this paper, to address these questions, we propose SQUARE, \textbf{S}emantic \textbf{Qu}ery-\textbf{A}ugmented Fusion and Efficient Batch \textbf{Re}ranking, a novel training-free framework for ZS-CIR. 
As illustrated in Figure~\ref{fig:method}, SQUARE operates in two stages. First, to enhance the embedding-based composed queries, we introduce Semantic Query-Augmented Fusion (SQAF) that incorporate high-level semantics. 
Build upon the weighted query fusion strategy in our prior work, WeiMoCIR~\cite{wu-taai24-WeiMoCIR}, we extend its training-free design by incorporating MLLMs for semantic enrichment. MLLMs are capable of generating rich, high-level descriptions that capture nuanced relationships between the reference image and modification text. The embedding-based query is then enhanced with these semantic cues to strengthen the query-target alignment. This leads to more accurate global retrieval and is particularly beneficial when using lightweight VLMs with limited capacity.

Second, we introduce \textbf{E}fficient \textbf{B}atch \textbf{R}eranking (EBR) that further exploits MLLM's multimodal reasoning for reranking. 
Building on the findings that visual marks on image regions enhance the fine-grained visual grounding abilities of MLLMs~\cite{yang-arxiv23-SoM,koh-acl24-VisualWebArena}, we annotate each candidate image with a distinct label and a bounding box, and then arrange them into a single grid image. 
This holistic view enables the MLLM to perform joint comparison of all candidates within a single forward pass, leading to both faster inference and more accurate reranking.
By encouraging the model to reason about inter-image differences rather than evaluating each image in isolation, EBR promotes richer comparative understanding.
Furthermore, it is designed as a standalone module that can be seamlessly integrated into existing retrieval pipelines.

Our method adopts a modular, coarse-to-fine strategy, offering high flexibility in model selection based on the desired trade-offs between efficiency and effectiveness. Moreover, it can seamlessly integrate different VLMs and MLLMs without additional training, making it adaptable to diverse computational budgets and performance requirements.
In summary, our main contributions are as follows:
\begin{itemize}
\item We propose SQUARE, a novel training-free framework for ZS-CIR, that follows a coarse-to-fine pipeline. SQUARE enables effective retrieval without requiring model fine-tuning or annotated triplets.
\item Our SQAF enhances the embedding-based image-text fusion with MLLM-generated descriptions of the target image. 
This straightforward approach not only boosts retrieval performance, but also enhances the interpretability of the composed query.
\item By harnessing MLLMs’ multimodal reasoning to compare top candidates in a single forward pass, our EBR offers an efficient reranking strategy that significantly improves result quality while incurring minimal computational overhead.
\item Extensive experiments on four CIR benchmarks show that SQUARE consistently achieves strong performance. Notably, it remains effective even with lightweight backbone models, demonstrating both practicality and scalability.
\end{itemize}

\section{Related Work} 

\subsection{Zero-Shot Composed Image Retrieval}
Zero-shot composed image retrieval (ZS-CIR) aims to retrieve images that match a reference image modified by natural language instructions, without requiring additional task-specific training.
Most existing ZS-CIR methods build on textual inversion, where abundant image–text pairs are used to train a mapping network that transforms a reference image into a pseudo-word token within CLIP’s token embedding space. The pseudo-word token is then concatenated with the modification text for retrieval.
The seminal work in this line, Pic2Word~\cite{saito-cvpr23-Pic2Word}, introduces a lightweight mapping network for this purpose. SEARLE~\cite{baldrati-iccv23-SEARLE} adopts a more sophisticated two-stage strategy: it first generates pseudo-word tokens for unlabeled images via optimization-based textual inversion with a GPT-guided regularization loss, and then distills this knowledge into a mapping network for efficient inference.
However, the pseudo-word tokens produced by these approaches primarily capture global visual information. 
To address this limitation, KEDs~\cite{suo-cvpr24-KEDs} further enhances the representation by leveraging an external image–caption database to enrich the pseudo-word tokens with fine-grained attribute details.
Since only a subset of visual features is relevant to the modification intent, Context-I2W~\cite{tang-aaai24-Context-I2W} employs an intent-view selector and a visual target extractor to learn a mapping network that adaptively attends to the context-dependent visual cues.
While these methods have made notable progress, they still depend on training mapping networks with large collections of image–text pairs. This limitation has motivated the development of training-free ZS-CIR approaches, which enable retrieval directly through pretrained models used as off-the-shelf tools.

\subsection{Training-Free Zero-shot Composed Image Retrieval}
Training-free ZS-CIR approaches typically build on the joint image–text semantic space established by VLMs, while also exploiting the advanced reasoning capabilities of LLMs and MLLMs. We categorize existing methods into three groups: methods relying solely on VLMs, methods combining VLMs with LLMs, and methods integrating VLMs with MLLMs.



\paragraph{VLM-only approaches}
VLMs like CLIP~\cite{radford-icml21-CLIP} and BLIP~\cite{li-icml22-BLIP} provide a pre-trained semantic space that aligns images and text, and they have been a fundamental component of recent CIR methods. Building on this joint embedding space, Slerp~\cite{jang-eccv24-Slerp}, the only training-free VLM-only approach, models the composed query as an intermediate representation between the embeddings of the reference image and modification text, obtained through spherical linear interpolation. 
Despite its simplicity, this interpolation strategy has demonstrated promising performance in ZS-CIR.



\paragraph{VLM+LLM approaches} 
Approaches in this category leverage the reasoning capabilities of LLMs to more effectively capture user intent in CIR.
LLMs excel at natural language reasoning and compositional understanding, making them well-suited for generating rich, human-understandable textual representations of the desired target image.
CIReVL~\cite{karthik-iclr24-CIReVL} reformulates CIR as a text-to-image retrieval task: it first employs a VLM to generate a caption for the reference image and then uses an LLM to integrate this caption with the modification text, producing a refined description of the target image for retrieval.
To address the limitation that a single generated caption may not fully capture the user intent, methods such as LDRE~\cite{yang-sigir24-LDRE} and SEIZE~\cite{yang-acmmm24-seize} expand the space of possible interpretations by generating multiple captions. The VLM provides diverse captions for the reference image, which are then refined by the LLM in light of the modification text, thereby covering the potential semantics of the desired target image.
However, since the reference image caption is produced by the VLM independently of the modification text, it may omit visual details that are crucial for the intended edits. This could result in descriptions that do not fully reflect the target image.
Although generating multiple diverse captions can partially mitigate this issue by broadening semantic coverage, the process is computationally expensive and risks introducing redundant or noisy descriptions.

\paragraph{VLM+MLLM approaches} 
The emergence of MLLMs, such as GPT-4o~\cite{openai-2024-gpt4-technical-report} and Gemini, has opened new avenues for ZS-CIR by enabling advanced reasoning over multimodal inputs, thereby addressing the limitations of VLMs+LLMs approaches. 
OSrCIR~\cite{tang-cvpr25-OSrCIR} leverages chain-of-thought (CoT) reasoning jointly interpret the visual content and the modification text, producing target descriptions that more closely align with the intended image in a single-stage reasoning process. 
To better preserve both the explicit modification and implicit visual cues, MCoT-RE~\cite{park-smc25-MCoT-RE} introduces multi-faceted CoT and generates two distinct captions: one highlighting the modification and the other capturing the broader visual–textual context. 
ImageScope~\cite{luo-acmwww25-ImageScope} captures the user intent at multiple semantic granularities to improve robustness.
In contrast, WeiMoCIR~\cite{wu-taai24-WeiMoCIR} takes a different perspective by using an MLLM to generate captions for the database images; during retrieval, it considers both query-to-image and query-to-caption similarities. 

Our work shares a similar spirit with previous studies by using an MLLM to generate target image descriptions; however, rather than directly using the captions for text-to-image retrieval~\cite{tang-cvpr25-OSrCIR}, we enhance the embedding-based query with semantic information from the MLLM-generated captions. 
This additional guidance yields a more informative query representation that better aligns with the intended target image.

\subsection{MLLMs for Reranking}
Beyond multimodal query composition, MLLMs have also been explored for reranking.
A common strategy is to prompt MLLMs with questions to assess whether specific aspects are present in candidate images. For instance, GRB~\cite{sun-arxiv23-GRB} verifies the existence of local concepts by posing binary (Yes/No) questions to an MLLM.
Based on such binary judgments, MM-EMBED~\cite{lin-iclr25-MMEMBED} further derives relevance scores to enable finer-grained reranking.
Rather than performing a single refinement, ImageScope~\cite{luo-acmwww25-ImageScope} adopts a multi-stage refinement process: predicate propositions are first verified locally for each candidate image, followed by global pairwise comparisons between the reference image and the top-ranked candidates, with both stages relying on binary decisions.

In contrast to the existing approaches, which rely on binary verification of individual images, we exploit the multimodal reasoning capabilities of MLLMs to rerank candidates by jointly comparing all images with the aid of visual marks. This reranking strategy can be performed in a single forward pass, enabling faster inference.

\section{Methodology}
CIR is a multimodal retrieval task in which a query consists of a reference image $I_r$ and a modification text $T_m$.
The goal is to retrieve a target image $I_t$ from a gallery of images $\mathcal{D}=\{I_i\}_{i=1}^{N} $ that captures the visual characteristics of $I_r$ while reflecting the semantic changes described in $T_m$. 
In training-free ZS-CIR, no task-specific training is required. Instead, pretrained VLMs, such as CLIP, are typically leveraged to perform the retrieval directly.
Specifically, the reference image and the text modifier are encoded into a joint embedding space using the VLM's image encoder $\mathcal{E}_{\text{img}}(\cdot)$ and text encoder $\mathcal{E}_{\text{txt}}(\cdot)$.
A composition function, \( f(\mathcal{E}_{\text{img}}(I_r), \mathcal{E}_{\text{txt}}(T_m) )\), then fuses these embeddings into a single query representation.
Retrieval is then performed by ranking the gallery images based on their similarity (\eg, cosine similarity) to the composed query embedding. 
This approach enables CIR without additional training by relying on the pretrained alignment between visual and textual modalities in VLMs.

To improve retrieval accuracy in this training-free setting, we introduce SQUARE, which is driven by our core idea of leveraging an MLLM $\mathcal{M}$ to enhance both the composed query and the candidate reranking.
As illustrated in Figure~\ref{fig:method}, our SQAF module employs the MLLM to generate a target image caption by reasoning over the user's intent expressed in the reference image $I_r$ and the modification text $T_m $. This caption is then incorporated into the composed query representation boost global ranking performance.
In our EBR module, the MLLM is further utilized to rerank a shortlist of the top candidate images by comparing their visual content against the reference image and modification text.

\subsection{SQAF: Semantic Query-Augmented Fusion for Richer Composed Queries}
In a training-free setting with VLMs, a common approach to construct the query representation is to perform a weighted fusion of the visual embedding of the reference image and the textual embedding of the modification text:
\begin{equation}\label{eqn:init-fusion}
\mathbf{q}_{\text{vlm}} = (1 - \alpha)\cdot \mathcal{E}_{\text{img}}(I_r) + \alpha \cdot \mathcal{E}_{\text{txt}}(T_m),
\end{equation}
where $ \mathbf{q}_{\text{vlm}} \in \mathbb{R}^d$ is the resulting $d$-dimensional VLM-based query representation, and $\alpha \in [0,1]$ is a hyperparameter that controls the relative contributions of the image and text.
However, while this fusion helps the query reflect both the visual appearance and the desired change, it is insufficient for a truly comprehensive understanding of the user's intent. This is mainly because the reference image, while visually rich, is often semantically ambiguous, and the modification text, while semantically clear, lacks detailed visual information. To address this limitation, we leverage the MLLM's powerful multimodal reasoning ability to enrich the query, thereby enabling a more complete capture of the user's complex compositional request.

\begin{figure}[tb]
    \centering
    \includegraphics[width=.7\textwidth]{figures/dfs_prompt.pdf}
    \caption{Example prompt used for generating the imagined target image caption. Our prompt is composed of three handcrafted few-shot examples and a set of explicit rules that guide the MLLM to generate a concise and concrete description of the target image based on a given reference image and textual modification.}
    \Description{Prompt for the Semantic Query-Augmented Fusion stage, consisting of three in-context examples and rule-based instructions. The MLLM is prompted to describe the expected target image directly, using simple, visual language, without abstract terms or proper nouns.}
    \label{fig:dfs_prompt}
\end{figure}

\paragraph{MLLM-based Query Augmentation}
Given their powerful multimodal reasoning capabilities, MLLMs are suited to infer a user's intent from complex, multi-modal queries. We leverage this ability to generate a high-level description of the desired target image. 
As depicted in the upper part of Figure~\ref{fig:method}, the MLLM $\mathcal{M}_1$ takes as input the reference image $I_r$, the text modifier $T_m$, and a reflective prompt $P_{\text{caption}}$. 
This prompt, whose structure is shown in Figure~\ref{fig:dfs_prompt}, guides the MLLM's reasoning with a few in-context examples. By demonstrating the desired task and output format, these examples instruct the MLLM to articulate the desired compositional changes. 
The generation of the target image description $T_t$ is formalized as follows:
\begin{equation}
T_t = \mathcal{M}_1(I_r, T_m, P_{\text{caption}}).
\end{equation}
The resulting $T_t$ serves as a human-readable interpretation of the user's intent, grounded in a multimodal understanding of the input. Unlike an opaque vector fusion, this caption is inherently interpretable and functions as a text-based proxy for the target image by explicitly articulating the compositional request.

\paragraph{Final Query Construction}
In the final stage of query construction, we integrate the semantic strengths of the MLLM with the representational power of the VLM.
While the VLM-based query $\mathbf{q}_{\text{vlm}}$ encodes both visual and textual features, it may fall short in explicitly capturing high-level, compositional semantics.
On the other hand, the MLLM-generated caption $T_t$ provides a human-readable, semantically rich description of the target image by reasoning over the reference image $I_r$ and the modification text $T_m$.
We embedding this caption using the VLM's text encoder $\mathcal{E}_{\text{txt}}(\cdot)$ and fuse it with $\mathbf{q}_{\text{vlm}}$ to obtain the final query:
\begin{equation}\label{eqn:final-fusion}
\mathbf{q} = (1 - \beta) \cdot \mathbf{q}_{\text{vlm}} + \beta \cdot \mathcal{E}_{\text{txt}}(T_t),
\end{equation}
where $\beta \in [0,1]$ is a hyperparameter that controls the influence of the MLLM's semantic description.
By appropriately setting $\beta$, the fusion integrates the multimodal alignment of the VLM with the high-level reasoning of the MLLM. This allows the expanded intent provided by the MLLM captions to enrich semantics while still being constrained by the information in the VLM-based query, ultimately yielding a representation that is both semantically richer and more closely aligned with the user’s intent.

\paragraph{Global Ranking}
We first encode all gallery images $I_i \in \mathcal{D}$ into image embeddings using the VLM's image encoder $\mathcal{E}_{\text{img}}(\cdot)$.
For each gallery image, we compute a similarity score $S_i$ by calculating the cosine similarity between the final query representation $\mathbf{q}$ and its image embedding $\mathcal{E}_{\text{img}}(I_i)$:
\begin{equation}
S_i = \frac{\mathbf{q}\cdot \mathcal{E}_{\text{img}}(I_i)}{ \Vert\mathbf{q} \Vert \Vert \mathcal{E}_{\text{img}}(I_i)\Vert}, \quad \forall I_i \in \mathcal{D}.
\end{equation}
Given these scores, we select the top-$K$ most relevant images to the query, forming an ordered list of candidate images:
\begin{equation}
\mathcal{C}_K = (I^{(0)}, I^{(1)}, \dots, I^{(K-1)}),
\end{equation}
where the superscript denotes the rank position in descending order of similarity.
The candidate set $\mathcal{C}_K$ is then passed to the reranking stage for finer-grained alignment with the user's intent.

\subsection{EBR: Efficient Batch Reranking for Finer-grained Alignment}
\begin{figure}[tb]
    \centering
    \includegraphics[width=.7\textwidth]{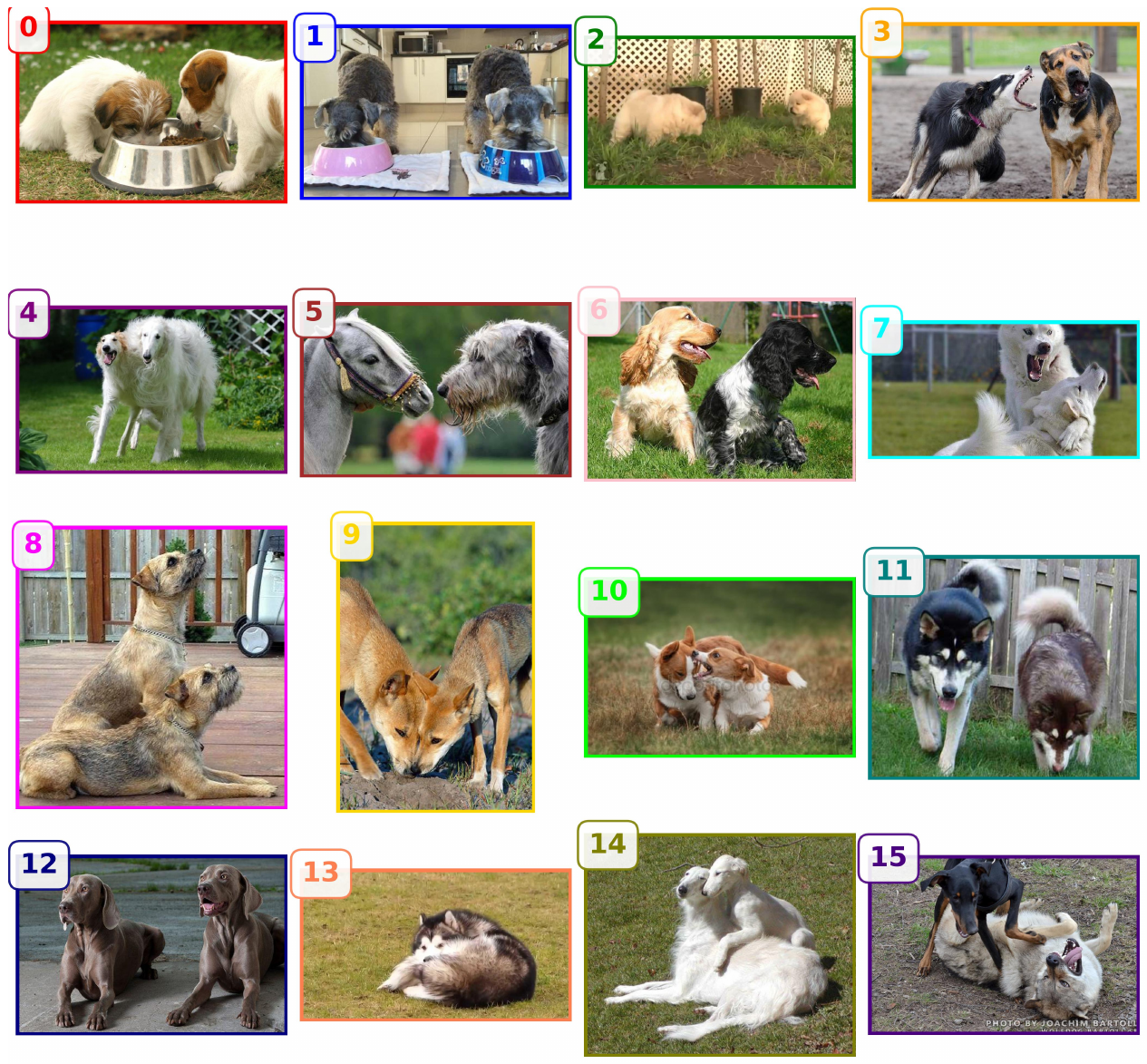}
    \caption{An example grid image used in the reranking process. Each candidate image is marked with a colored bounding box and a numeric label at the top-left corner. These identifiers enable the MLLM to reference specific images during reasoning and generate an updated ranking based on the image relevance to the user's intent.}
    \Description{Each candidate image is marked with a color-coded bounding box and a number in the top-left corner. The MLLM uses these labels to reason across the image grid and returns a new ranking based on the visual-textual match.}
    \label{fig:rerank_grid_image}
\end{figure}

Although our composed query, which fuses the VLM-based representation with the MLLM-generated caption, provides a decent coarse-level alignment with candidate images, it may still fall short in capturing subtle compositional relationships and fine-grained semantic cues.
Inspired by SoM~\cite{yang-arxiv23-SoM}, our intuition is that an MLLM, with its ability to jointly reason over visual and textual inputs, can evaluate all candidates in a shared context, compare them with the user’s intent, and produce a more semantically accurate ordering.

To prepare for reranking, we annotate each candidate with a colored bounding box and a numeric label at the top-left corner.
In our implementation, we set \(K = M^2\) and arrange these annotated images into a compact $M \times M$ grid image $\mathcal{G}$. 
Formally,
we define a grid construction function \(\mathrm{Grid}(\cdot)\) that arranges the inputs row-wise into an \(M\times M\) layout:  
\begin{equation}
\mathcal{G} = \mathrm{Grid}\big(\{\mathrm{Annotate}(I^{(i)},\, i)\}_{i=0}^{K-1};\ \text{rows}=M, \text{cols}=M\big),
\end{equation}
where \(\mathrm{Annotate}(I,\, i)\) overlays the image \(I\) with a bounding box and its rank index \(i\) for explicit referencing.
Figure~\ref{fig:rerank_grid_image} shows an example with $M=4$.
These visual and numeric identifiers enable the MLLM to reference specific images during multimodal reasoning and to produce an updated ranking that reflects each candidate’s relevance to the user’s intent. 

\begin{figure}[tb]
    \centering
    \includegraphics[width=.7\textwidth]{figures/rerank_prompt_text.pdf}
    \caption{Example prompt for the reranking stage. This prompt provides the MLLM with the reference image, the textual modification, and a grid of candidate images. The MLLM's task is to reason over the visual differences and produce an updated ranking based on the candidates' alignment with the user's intent.}
    \Description{Text prompt showing reference image, textual modification, candidate image grid, and instructions to the MLLM for reranking.}
    \label{fig:rerank_prompt_text}
\end{figure}


With the constructed grid image, a critical design decision in our reranking stage concerns how to best represent the user's intent to the MLLM.
Although one could use the target image caption from the SQAF module as a proxy for the user's intent, we instead choose to use the reference image and the modification text directly. This choice offers two advantages:
(1) It allows our EBR to function as a standalone reranking module, independent of the method used to obtain the top-$K$ candidate set, and can thus be seamlessly integrated into any existing CIR pipeline for refinement.
(2) Directly providing the reference image and the modification text allows the MLLM to perform fresh multimodal reasoning over the original inputs, reducing the potential information loss or bias that might arise from relying solely on the caption. 

Based on this design choice, the MLLM $\mathcal{M}_2$ takes as input the reference image $I_r$, the modification text $T_m$, the grid image $G$, and the reranking prompt $P_{\text{rerank}}$. It then returns an ordered, updated index sequence $\pi'$, defined as:
\begin{equation}
\pi' = \mathcal{M}_2 (I_r, T_m, \mathcal{G}, P_{\text{rerank}}), 
\end{equation}
where the elements of $\pi'$ are drawn from $\{0, 1, \dots, K-1\}$, and the structure of the reranking prompt is shown in
Figure~\ref{fig:rerank_prompt_text}. While the prompt requests that the MLLM consider all candidate images, $\pi'$ may contain only a partial list of indices. This implies that the model deems only a subset of the candidates relevant. In such cases, we preserve the initial order of the remaining candidates and append them to the end of the MLLM’s ranking to produce a complete ordering. 
That is, the final ranking $\pi = \pi' \oplus \pi_{\text{rem}}$, where $\oplus$ denotes the sequence concatenation, and $\pi_{\text{rem}}$ is the sequence of indices in $\{0, 1, \dots, K-1\} \setminus \text{Set}(\pi')$ for the remaining candidates, sorted in their initial order. 
The final ordered list of ranked images $\mathcal{R}$ is obtained by applying this final index sequence to the original set of candidate images:
\begin{equation}
\mathcal{R} = (I^{(\pi_0)}, I^{(\pi_1)}, \dots, I^{(\pi_{K-1})}).
\end{equation}

\section{Experiments}

\subsection{Datasets and Setup}
\paragraph{Datasets} 
We evaluate SQUARE on four public CIR benchmarks, namely CIRR~\cite{liu-iccv21-CIRR}, CIRCO~\cite{baldrati-iccv23-SEARLE}, FashionIQ~\cite{wu-cvpr21-FashionIQ}, and GeneCIS~\cite{vaze-cvpr23-genecis}.
\textbf{CIRR}~\cite{liu-iccv21-CIRR}, built upon NLVR2~\cite{suhr-neurips19-NLVR2}, consists of real-world images with diverse visual content. The modification text typically involves changes at object- or scene-level. We report results on the test set evaluation server\footnote{\url{https://cirr.cecs.anu.edu.au/test_process/}}. 
\textbf{CIRCO} is a larger benchmark built from natural images sourced from the COCO 2017 unlabeled set~\cite{lin-eccv14-COCO}. Each query is linked with multiple ground-truths images, averaging 4.53 per query.
\textcolor{black}{With a retrieval set comprising 123,403 images, the dataset presents a greater number of distractors, making the task significantly more challenging.} We report results on the test set evaluation server\footnote{\url{https://circo.micc.unifi.it/evaluation/}}.
\textbf{FashionIQ}~\cite{wu-cvpr21-FashionIQ} is a well-known benchmark for evaluating CIR in the fasion domain. It comprises real-world product images across three categories: \textit{Dress}, \textit{Shirt}, and \textit{Toptee}.
Each query comprises a reference image and a natural language modification describing subtle appearance changes, such as color, length, or style, supporting fine-grained compositional reasoning in retrieval.
We follow previous studies~\cite{tang-cvpr25-OSrCIR,yang-sigir24-LDRE} and report results on the validation set.
\textbf{GeneCIS} is built from MS-COCO~\cite{lin-eccv14-COCO} and Visual Attributes in the Wild (VAW)~\cite{pham-cvpr21-VAW}, focusing on four CIR task types: modifying or focusing on specific attributes or objects. It includes 8,032 query pairs, each paired with a small retrieval set (typically 15 images, or 10 for the "Focus on an Attribute" task), enabling controlled, fine-grained evaluation of semantic transformations in retrieval.

\paragraph{Evaluation metrics} 
We report Recall@K (R@K) with K = 1, 5, and 10 for CIRR, which evaluates whether the ground-truth image appears among the top-K retrieved results. Additionally, we report Recall$_\text{Subset}$@K with K = 1, 2, and 3 in the subset setting of CIRR, 
where the retrieval is performed on a restricted candidate pool composed of images that are semantically similar to the query, especially including challenging negatives selected to prevent trivial discrimination.
For CIRCO, we adopt the mean Average Precision at K (mAP@K) with K = 5, 10, 25, and 50, which accounts for multiple ground-truths per query and provides a more fine-grained evaluation of ranked retrieval performance.
For FashionIQ, we compute R@10 and R@50 separately for each clothing category and report the average across categories.
In the case of GeneCIS, we evaluate R@K (K = 1, 2, 3) on four distinct tasks, each designed to test the model's ability to handle different types of semantic modification.

\paragraph{Implementation Details} We use the CLIP models pretrained on the LAION-2B English subset of the LAION-5B dataset~\cite{scuhmann-neurips22-LAION-5B} as our VLM, denoted by their backbone and patch size (\eg, ViT-B/32). Specifically, we use the CLIP visual encoder to extract image features and the text encoder to embed both the user-provided modification texts and the MLLM-generated descriptions. 
We employ GPT-4o~\cite{openai-2024-gpt4-technical-report} as the MLLM, with the temperature set to its default value of 1.0.
For the SQAF stage, which produces an initial global ranking, we set the fusion weight $\alpha$ in Equation~\eqref{eqn:init-fusion} to 0.7 and $\beta$ in Equation~\eqref{eqn:init-fusion} to 0.6.
The top 16 retrieved images from SQAF are arranged into a $4 \times 4$ grid image and passed to the reranking module. All experiments are c
All experiments are conducted on a single NVIDIA RTX 3090 GPU using Python 3.8 (Conda) and PyTorch 2.3.0.

\subsection{Main Results}

\begin{table}[tb]
  \caption{Comparison of retrieval performance on the CIRR and CIRCO test sets. Best results are highlighted in bold. Our method consistently achieves superior performance across all metrics and CLIP backbones.}
  \label{tab:cirr_circo_performance}
  \centering
  \setlength{\tabcolsep}{4pt}
  \begin{tabular*}{\columnwidth}{@{} @{\extracolsep{\fill} } ll *{10}{c} @{} }
    \toprule
    &
    & \multicolumn{6}{c}{CIRR} 
    & \multicolumn{4}{c}{CIRCO}\\
    \cmidrule(lr){3-8}\cmidrule(lr){9-12}
    \multicolumn{2}{c}{Metric}
    & \multicolumn{3}{c}{Recall@K} 
    & \multicolumn{3}{c}{Recall$_\text{Subset}$@K} 
    & \multicolumn{4}{c}{mAP@K}\\
    \cmidrule(lr){3-5} \cmidrule(lr){6-8} \cmidrule(lr){9-12}
    Backbone  & Method                                                   & K=1            & K=5            & K=10           
    & K=1            & K=2            & K=3            & K=5            & K=10           & K=25    & K=50 \\
    \midrule

    \multirow{8}{*}{ViT-B/32}
    & Slerp~\cite{jang-eccv24-Slerp} {\footnotesize (ECCV'24)}          & 24.22          & 50.94          & 63.49          
    & 57.86          & 78.27          & 89.25          & 6.35           & 7.11           & 8.12           & 8.75      \\
    & CIReVL~\cite{karthik-iclr24-CIReVL} {\footnotesize (ICLR'24)}     & 23.94          & 52.51          & 66.00          
    & 60.17          & 80.05          & 90.19          & 14.94          & 15.42          & 17.00          & 17.82      \\
    & LDRE~\cite{yang-sigir24-LDRE} {\footnotesize (SIGIR'24)}          & 25.69          & 55.13          & 69.04          
    & 60.53          & 80.65          & 90.70          & 17.96          & 18.32          & 20.21          & 21.11      \\
    & SEIZE~\cite{yang-acmmm24-seize} {\footnotesize (ACM MM'24)}       & 27.47          & 57.42          & 70.17          
    & 65.59          & 84.48          & 92.77          & 19.04          & 19.64          & 21.55          & 22.49      \\
    & OSrCIR~\cite{tang-cvpr25-OSrCIR}  {\footnotesize (CVPR'25)}       & 25.42          & 54.54          & 68.19          
    & 62.31          & 80.86          & 91.13          & 18.04          & 19.17          & 20.94          & 21.85      \\
    & ImageScope~\cite{luo-acmwww25-ImageScope} {\footnotesize (WWW'25)}& 38.43          & 66.27          & 76.96          
    & 75.93          & 89.21          & 94.63          & 25.26          & 25.82          & 27.15          & 28.11      \\
    & \textbf{SQUARE w/o EBR} {\footnotesize (Ours)}                    & 33.49          & 65.57          & 76.72          
    & 63.74          & 82.39          & 91.57          & 20.89          & 21.47          & 23.38          & 24.31      \\ 
    & \textbf{SQUARE} {\footnotesize (Ours)}                            & \textbf{45.04} & \textbf{72.05} & \textbf{79.95} 
    & \textbf{80.94} & \textbf{93.13} & \textbf{97.23} & \textbf{28.95} & \textbf{28.46} & \textbf{29.66} & \textbf{30.59} \\ 
    \midrule

    \multirow{9}{*}{ViT-L/14}
    & Slerp~\cite{jang-eccv24-Slerp} {\footnotesize (ECCV'24)}          & 24.43          & 49.93          & 62.29          
    & 57.71          & 77.59          & 88.80          & 8.76           & 9.84           & 11.30          & 11.99      \\
    & CIReVL~\cite{karthik-iclr24-CIReVL} {\footnotesize (ICLR'24)}     & 24.55          & 52.31          & 64.92          
    & 59.54          & 79.88          & 89.69          & 18.57          & 19.01          & 20.89          & 21.80      \\
    & LDRE~\cite{yang-sigir24-LDRE}  {\footnotesize (SIGIR'24)}         & 26.53          & 55.57          & 67.54          
    & 60.43          & 80.31          & 89.90          & 23.35          & 24.03          & 26.44          & 27.50      \\
    & SEIZE~\cite{yang-acmmm24-seize} {\footnotesize (ACM MM'24)}       & 28.65          & 57.16          & 69.23          
    & 66.22          & 84.05          & 92.34          & 24.98          & 25.82          & 28.24          & 29.35      \\
    & OSrCIR~\cite{tang-cvpr25-OSrCIR}  {\footnotesize (CVPR'25)}       & 29.45          & 57.68          & 69.86          
    & 62.12          & 81.92          & 91.10          & 23.87          & 25.33          & 27.84          & 28.97      \\
    & ImageScope~\cite{luo-acmwww25-ImageScope} {\footnotesize (WWW'25)}& 39.37          & 67.54          & 78.05          
    & 76.36          & 89.40          & 95.21          & 28.36          & 29.23          & 30.81          & 31.88      \\
    & MCoT-RE~\cite{park-smc25-MCoT-RE}  {\footnotesize (IEEE SMC'25)}  & 39.52          & 69.18          & 79.28          
    & 70.19          & 86.34          & 93.83          & --             & --             & --             & --         \\
    & \textbf{SQUARE w/o EBR} {\footnotesize (Ours)}                    & 36.70          & 67.54          & 78.19          
    & 66.41          & 83.81          & 92.34          & 26.74          & 27.47          & 29.79          & 30.95      \\  
    & \textbf{SQUARE} {\footnotesize (Ours)}                            & \textbf{44.94} & \textbf{72.29} & \textbf{80.65} 
    & \textbf{80.55} & \textbf{93.30} & \textbf{97.23} & \textbf{33.94} & \textbf{33.79} & \textbf{35.41} & \textbf{36.57}\\ 

    \midrule

    \multirow{7}{*}{ViT-G/14}
    & CIReVL~\cite{karthik-iclr24-CIReVL} {\footnotesize (ICLR'24)}     & 34.65          & 64.29          & 75.06          
    & 67.95          & 84.87          & 93.21          & 26.77          & 27.59          & 29.96          & 31.03      \\
    & LDRE~\cite{yang-sigir24-LDRE}  {\footnotesize (ICLR'24)}          & 36.15          & 66.39          & 77.25          
    & 68.82          & 85.66          & 93.76          & 31.12          & 32.24          & 34.95          & 36.03      \\
    & SEIZE~\cite{yang-acmmm24-seize}  {\footnotesize (ACM MM'24)}      & 38.87          & 69.42          & 79.42          
    & 74.15          & 89.23          & 95.71          & 32.46          & 33.77          & 36.46          & 37.55      \\
    & OSrCIR~\cite{tang-cvpr25-OSrCIR} {\footnotesize (CVPR'25)}        & 37.26          & 67.25          & 77.33          
    & 69.22          & 85.28          & 93.55          & 30.47          & 31.14          & 35.03          & 36.59      \\
    & MCoT-RE~\cite{park-smc25-MCoT-RE} {\footnotesize (IEEE SMC'25)}   & 39.37          & 68.92          & 79.30          
    & 70.82          & 86.92          & 93.88          & --             & --             & --             & --         \\
    & \textbf{SQUARE w/o EBR} {\footnotesize (Ours)}                    & 38.72          & 69.47          & 79.30          
    & 68.10          & 85.37          & 93.37          & 30.89          & 32.06          & 34.52          & 35.64      \\  
    & \textbf{SQUARE} {\footnotesize (Ours)}                            & \textbf{45.13} & \textbf{72.60} & \textbf{80.96} 
    & \textbf{79.90} & \textbf{92.94} & \textbf{97.06} & \textbf{35.61} & \textbf{35.64} & \textbf{37.70} & \textbf{38.82} \\
    \bottomrule
  \end{tabular*}
\end{table}


We compare SQUARE against representative training-free ZS-CIR methods across three categories: VLM-only, represented by Slerp~\cite{jang-eccv24-Slerp}; VLM+LLM, including CIReVL~\cite{karthik-iclr24-CIReVL}, LDRE~\cite{yang-sigir24-LDRE}, and SEIZE~\cite{yang-acmmm24-seize}; and VLM+MLLM, comprising OSrCIR~\cite{tang-cvpr25-OSrCIR}, ImageScope~\cite{luo-acmwww25-ImageScope}, and MCoT-RE~\cite{park-smc25-MCoT-RE}.

\paragraph{CIRR results}
On CIRR, as shown in the left part of Table~\ref{tab:cirr_circo_performance}, VLM+LLM approaches generally show improved performance compared to VLM-only approaches, indicating that the high-level semantic information provided by LLMs helps better capture the modification intent.
We also observed that the performance of the VLM+LLM methods improves with larger VLP models. This highlights the importance of high-capacity CLIP representations. 
In contrast, VLM+MLLM-based methods, benefiting from the strong reasoning and semantic understanding capabilities of MLLMs, typically outperform other approaches that do not use MLLMs. 
Our method, when using only the SQAF module (denoted as SQUARE w/o EBR), outperforms OSrCIR~\cite{tang-cvpr25-OSrCIR}  across nearly all settings and CLIP backbones, with the exception of ViT-G/14 on the CIRR subset.
When further employing the EBR module, SQUARE achieves the highest retrieval accuracy among all the methods compared.
While both ImageScope and MCoT-RE also adopt coarse-to-fine strategies, our approach is more efficient: the prompt requires only a small number of exemplar texts to identify global ranking candidates, avoiding the token-intensive CoT process.
A simple reranking of these top candidates then yields substantial gains, resulting in clear improvements over ImageScope and MCoT-RE.
These results highlight the effectiveness of SQUARE, which combines semantic query fusion and MLLM-guided reranking for training-free ZS-CIR.



\paragraph{CIRCO results} On CIRCO, as presented in the right part of Table~\ref{tab:cirr_circo_performance}, MLLM-based methods generally outperform non-MLLM-based ones when using smaller CLIP backbones. With larger backbones, however, non-MLLM approaches such as LDRE and SEIZE, which leverage an LLM to perform compositional reasoning over multiple captions generated for the reference image and the modification text, achieve competitive performance.
Our method, SQUARE, achieves consistently superior performance across all CLIP backbones. This improvement is largely attributed to the effectiveness of the proposed reranking strategy, which yields an average boost of 6.66\% in mAP@5.

\paragraph{FashionIQ results} 
Table~\ref{tab:fashioniq_performance} presents the results on the FashionIQ validation set for the fashion attribute manipulation task. Our SQUARE w/o EBR already outperforms all other methods in both R@10 and R@50 metrics with the ViT-B/32 and ViT-L/14 backbones. 
This demonstrates the ability of our proposed SQAF query composition to retrieve semantically relevant images.
Incorporating the EBR module further boosts R@10 by reranking only the top 16 candidate images. With the ViT-G/14 backbone, SQUARE w/o EBR and SQUARE achieve higher R@10 scores than SEIZE, although they slightly underperform in R@50. This suggests that our method is particularly effective in retrieving highly relevant images within the top-ranked results, achieving competitive results even without relying on multiple captions like SEIZE. 
Overall, SQUARE remains highly competitive in different settings.

\begin{table}[tb]
  \caption{Comparison of retrieval performance on the FashionIQ validation set. Best results are highlighted in bold. On average, our method outperforms all baselines in the R@10 metric across different backbones.}
  \label{tab:fashioniq_performance}
  \centering
  \begin{tabular*}{\columnwidth}{@{} @{\extracolsep{\fill} } ll *{8}{c} @{} }
    \toprule
    \multirow{1}{*}{} & \multirow{1}{*}{}  & \multicolumn{2}{c}{Shirt} & \multicolumn{2}{c}{Dress} & \multicolumn{2}{c}{Toptee} & \multicolumn{2}{c}{Average} \\
    \cmidrule(lr){3-4} \cmidrule(lr){5-6} \cmidrule(lr){7-8} \cmidrule(lr){9-10}
    Backbone & Method                                                     & R@10           & R@50           & R@10           & R@50           & R@10           & R@50           & R@10           & R@50           \\
    \midrule
    \multirow{8}{*}{ViT-B/32}
    & Slerp~\cite{jang-eccv24-Slerp} {\footnotesize (ECCV'24)}            & 23.75          &  40.92         & 20.53          & 41.00          & 26.98          & 46.77          & 23.75          & 42.90          \\
    & CIReVL~\cite{karthik-iclr24-CIReVL} {\footnotesize (ICLR'24)}       & 28.36          & 47.84          & 25.29          & 46.36          & 31.21          & 53.85          & 28.29          & 49.35          \\
    & LDRE~\cite{yang-sigir24-LDRE} {\footnotesize (SIGIR'24)}            & 27.38          & 46.27          & 19.97          & 41.84          & 27.07          & 48.78          & 24.81          & 45.63          \\
    & SEIZE~\cite{yang-acmmm24-seize}  {\footnotesize (ACM MM'24)}        & 29.38          & 47.97          & 25.37          & 46.84          & 32.07          & 54.78          & 28.94          & 49.86          \\
    & OSrCIR~\cite{tang-cvpr25-OSrCIR}  {\footnotesize (CVPR'25)}         & 31.16          & 51.13          & 29.35          & 50.37          & 36.51          & 58.71          & 32.34          & 53.40          \\
    & ImageScope~\cite{luo-acmwww25-ImageScope} {\footnotesize (WWW'25)}  & 31.65          & 50.15          & 26.82          & 46.31          & 35.80          & 55.94          & 31.42          & 50.80          \\
    & \textbf{SQUARE w/o EBR} {\footnotesize (Ours)}                      & 36.46          & \textbf{56.97} & 36.44          & \textbf{57.46} & 44.72          & \textbf{65.02} & 39.21          & \textbf{59.82} \\ 
    & \textbf{SQUARE} {\footnotesize (Ours)}                              & \textbf{38.37} & \textbf{56.97} & \textbf{36.99} & \textbf{57.46} & \textbf{46.40} & \textbf{65.02} & \textbf{40.59} & \textbf{59.82} \\
    \midrule
    \multirow{9}{*}{ViT-L/14}
    & Slerp~\cite{jang-eccv24-Slerp} {\footnotesize (ECCV'24)}            & 28.66          & 43.96          & 21.96          & 41.55          & 30.24          & 48.34          & 26.95          & 44.62          \\
    & CIReVL~\cite{karthik-iclr24-CIReVL}  {\footnotesize (ICLR'24)}      & 29.49          & 47.40          & 24.79          & 44.76          & 31.36          & 53.65          & 28.55          & 48.57          \\
    & LDRE~\cite{yang-sigir24-LDRE}  {\footnotesize (SIGIR'24)}           & 31.04          & 51.22          & 22.93          & 46.76          & 31.57          & 53.64          & 28.51          & 50.54          \\
    & SEIZE~\cite{yang-acmmm24-seize}   {\footnotesize (ACM MM'24)}       & 33.04          & 53.22          & 30.93          & 50.76          & 35.57          & 58.64          & 33.18          & 54.21          \\
    & OSrCIR~\cite{tang-cvpr25-OSrCIR} {\footnotesize (CVPR'25)}          & 33.17          & 52.03          & 29.70          & 51.81          & 36.92          & 59.27          & 33.26          & 54.37          \\
    & ImageScope~\cite{luo-acmwww25-ImageScope} {\footnotesize (WWW'25)}  & 32.87          & 51.07          & 26.17          & 46.15          & 35.03          & 55.12          & 31.36          & 50.78          \\
    & MCoT-RE~\cite{park-smc25-MCoT-RE} {\footnotesize (IEEE SMC'25)}     & 36.60          & 53.09          & 35.94          & 55.82          & 43.55          & 64.30          & 38.70          & 57.74          \\
    & \textbf{SQUARE w/o EBR} {\footnotesize (Ours)}                      & 39.79          & \textbf{59.22} & 35.99          & \textbf{57.51} & 45.54          & \textbf{64.97} & 40.44          & \textbf{60.57} \\ 
    & \textbf{SQUARE} {\footnotesize (Ours)}                              & \textbf{41.32} & \textbf{59.22} & \textbf{37.98} & \textbf{57.51} & \textbf{46.61} & \textbf{64.97} & \textbf{41.97} & \textbf{60.57} \\ 
    \midrule
    \multirow{7}{*}{ViT-G/14}
    & CIReVL~\cite{karthik-iclr24-CIReVL} {\footnotesize (ICLR'24)}       & 33.71          & 51.42          & 27.07          & 49.53          & 35.80          & 56.14          & 32.19          & 52.36          \\
    & LDRE~\cite{yang-sigir24-LDRE}  {\footnotesize (SIGIR'24)}           & 35.94          & 58.58          & 26.11          & 51.12          & 35.42          & 56.67          & 32.49          & 55.46          \\
    & SEIZE~\cite{yang-acmmm24-seize}  {\footnotesize (ACM MM'24)}        & 43.60          & \textbf{65.42} & \textbf{39.61} & \textbf{61.02} & 45.94          & \textbf{71.12} & 43.05          & \textbf{65.85} \\
    & OSrCIR~\cite{tang-cvpr25-OSrCIR}  {\footnotesize (CVPR'25)}         & 38.65          & 54.71          & 33.02          & 54.78          & 41.04          & 61.83          & 37.57          & 57.11          \\
    & MCoT-RE~\cite{park-smc25-MCoT-RE} {\footnotesize (IEEE SMC'25)}     & 42.35          & 59.81          & 34.51          & 56.67          & 45.74          & 67.57          & 40.87          & 61.35          \\
    & \textbf{SQUARE w/o EBR} {\footnotesize (Ours)}                      & 44.60          & 62.51          & 37.68          & 60.19          & 49.67          & 69.25          & 43.98          & 63.98          \\ 
    & \textbf{SQUARE} {\footnotesize (Ours)}                              & \textbf{45.04} & 62.51          & 37.68          & 60.19          & \textbf{49.87} & 69.25          & \textbf{44.20} & 63.98          \\ 
    \bottomrule
  \end{tabular*}
\end{table}

\paragraph{GeneCIS results}
Table~\ref{tab:genecis_performance} summarizes the retrieval results on GeneCIS. Our method, SQUARE, shows strong performance in most retrieval tasks, consistently achieving the highest R@1 scores with all backbones. 
However, for the change object task, SQUARE underperforms relative to other methods. 
We observe that in this task, the MLLM may overemphasize objects or background elements that are irrelevant to the search intent, thereby reducing overall ranking performance.
Despite this limitation, SQUARE remains highly competitive overall, especially in retrieving the most relevant images at the top ranks for diverse compositional queries.

\begin{table}[tb]
  \caption{Comparison of retrieval performance on GeneCIS. Best results are highlighted in bold. On average, our method achieves the highest R@1 scores across different tasks and backbones.}
  \label{tab:genecis_performance}
  \centering
  {\small \setlength{\tabcolsep}{3pt}
  \begin{tabular*}{\columnwidth}{@{} @{\extracolsep{\fill} } ll *{13}{c} @{} }
    \toprule
    & & \multicolumn{3}{c}{Focus Attribute}
      & \multicolumn{3}{c}{Change Attribute}
      & \multicolumn{3}{c}{Focus Object}
      & \multicolumn{3}{c}{Change Object}
      & \multicolumn{1}{c}{Average} \\
    \cmidrule(lr){3-5} \cmidrule(lr){6-8} \cmidrule(lr){9-11} \cmidrule(lr){12-14} \cmidrule(lr){15-15}
    Backbone & Method                     & R@1           & R@2           & R@3           & R@1           & R@2           & R@3           & R@1           & R@2           & R@3           & R@1           & R@2           & R@3           & R@1           \\
    \midrule
    \multirow{4}{*}{ViT-B/32}
    & CIReVL~\cite{karthik-iclr24-CIReVL} {\footnotesize (ICLR'24)} & 17.9          & 29.4          & 40.4          & 14.8          & 25.8          & 35.8          & 14.6          & 24.3          & 33.3          & 16.1          & 27.8          & 37.6          & 15.9          \\
    & OSrCIR~\cite{tang-cvpr25-OSrCIR} {\footnotesize (CVPR'25)}   & 19.4          & 32.7          & 42.8          & 16.4          & 27.7          & \textbf{38.1} & 15.7          & 25.7          & 35.8          & \textbf{18.2} & \textbf{30.1} & \textbf{39.4} & 17.4          \\
    & \textbf{SQUARE w/o EBR}   {\footnotesize (Ours)}                    & 20.3          & 33.8          & 44.6          & 14.0          & 24.9          & 34.6          & 15.6          & 26.9          & 37.6          & 15.6          & 26.8          & 37.4          & 16.4          \\
    & \textbf{SQUARE}  {\footnotesize (Ours)}               & \textbf{25.6} & \textbf{39.3} & \textbf{48.6} & \textbf{19.0} & \textbf{30.5} & 37.7          & \textbf{17.3} & \textbf{29.1} & \textbf{38.3} & 16.8          & 26.9          & 34.8          & \textbf{19.7} \\
    
    \midrule
    \multirow{4}{*}{ViT-L/14}
    & CIReVL~\cite{karthik-iclr24-CIReVL} {\footnotesize (ICLR'24)}& 19.5          & 31.8          & 42.0          & 14.4          & 26.0          & 35.2          & 12.3          & 21.8          & 30.5          & 17.2          & 28.9          & 37.6          & 15.9          \\
    & OSrCIR~\cite{tang-cvpr25-OSrCIR}  {\footnotesize (CVPR'25)}  & 20.9          & 33.1          & 44.5          & 17.2          & 28.5          & \textbf{37.9} & 15.0          & 23.6          & 34.2          & \textbf{18.4} & \textbf{30.6} & 38.3          & 17.9          \\
    & \textbf{SQUARE w/o EBR}  {\footnotesize (Ours)}                     & 20.3          & 34.0          & 44.2          & 14.4          & 25.6          & 35.1          & 17.3          & 27.5          & 38.0          & 17.2          & 29.1          & \textbf{38.7} & 17.3          \\
    & \textbf{SQUARE}  {\footnotesize (Ours)}               & \textbf{25.6} & \textbf{39.2} & \textbf{49.2} & \textbf{19.8} & \textbf{30.5} & \textbf{37.9} & \textbf{17.6} & \textbf{30.3} & \textbf{39.6} & 16.3          & 27.2          & 35.1          & \textbf{19.8} \\
    
    \midrule
    \multirow{4}{*}{ViT-G/14}
    & CIReVL~\cite{karthik-iclr24-CIReVL} {\footnotesize (ICLR'24)} & 20.5          & 34.0          & 44.5          & 16.1          & 28.6          & 39.4          & 14.7          & 25.2          & 33.0          & 18.1          & 31.2          & 41.0          & 17.4          \\
    & OSrCIR~\cite{tang-cvpr25-OSrCIR}  {\footnotesize (CVPR'25)}  & 22.7          & 36.4          & 47.0          & 17.9          & \textbf{30.8} & \textbf{42.0} & 16.9          & 28.4          & 36.7          & \textbf{21.0} & \textbf{33.4} & \textbf{44.2} & \textbf{19.6} \\
    & \textbf{SQUARE w/o EBR}  {\footnotesize (Ours)}                    & 21.6          & 35.6          & 44.9          & 14.8          & 26.8          & 37.1          & 16.8          & 28.4          & 38.8          & 15.7          & 28.9          & 38.6          & 17.2          \\
    & \textbf{SQUARE}   {\footnotesize (Ours)}              & \textbf{26.4} & \textbf{39.8} & \textbf{49.5} & \textbf{19.4} & 29.3          & 37.5          & \textbf{17.9} & \textbf{29.6} & \textbf{39.0} & 14.7          & 26.6          & 35.0          & \textbf{19.6} \\
    \bottomrule
  \end{tabular*}
  }
\end{table}

\subsection{Ablation Studies}
\label{sec:ablation}

\begin{table}[tb]
  \caption{Effect of different VLM backbones on the performance of the SQAF module (\ie, SQUARE w/o EBR). Retrieval results are reported on the CIRCO test set.}
  \label{tab:ablation_backbones_circo}
  \centering
  \begin{tabular*}{.6\columnwidth}{@{} @{\extracolsep{\fill} } l *{4}{c} @{} }
    \toprule
    Backbone            & mAP@5          & mAP@10         & mAP@25         & mAP@50         \\
    \midrule
    CLIP ViT-B/32       & 20.89          & 21.47          & 23.38          & 24.31          \\ 
    CLIP ViT-L/14       & 26.74          & 27.47          & 29.79          & 30.95          \\ 
    CLIP ViT-G/14       & \textbf{30.89} & \textbf{32.06} & \textbf{34.52} & \textbf{35.64} \\ 
    BLIP w/ ViT-B       & 25.41          & 26.34          & 28.72          & 29.68          \\ 
    BLIP w/ ViT-L       & 25.85          & 26.82          & 29.18          & 30.24          \\ 
    BLIP-2 w/ ViT-G     & 28.33          & 29.40          & 31.74          & 32.88 \\ 
    \bottomrule
  \end{tabular*}
\end{table}

The performance of SQUARE can be significantly influenced by the choice of VLMs and MLLMs, as well as the fusion hyperparameters. We analyze the impact of these factors to provide a deeper understanding of our method.

\paragraph{Effect of different VLMs on SQAF performance}
SQUARE relies on the SQAF module to generate a global ranking, and its retrieval quality is closely tied to the underlying VLMs.
Table~\ref{tab:ablation_backbones_circo} presents the impact of different VLM backbones on SQAF’s performance.
Among the evaluated models, CLIP ViT-G/14 achieves the highest mAP scores across all evaluated ranks (top-K), followed by CLIP ViT-L/14 and ViT-B/32, indicating that larger and more expressive backbones consistently yield better results. BLIP and BLIP-2 also demonstrate competitive performance, particularly BLIP-2 with ViT-G, which outperforms some CLIP variants at certain ranks. These findings highlight the flexibility of our retrieval framework and its ability to take advantage of advances in backbone architectures, with larger models offering stronger feature representations for composed image retrieval.

\paragraph{Effect of different MLLMs on SQAF performance}
The choice of MLLM within our SQAF module significantly impacts retrieval performance, as shown in Table~\ref{tab:ablation_mllms_circo_SQAF}. 
When the MLLM-generated description of the target image is excluded (indicated as ``w/o MLLM''), the mAP scores are substantially lower. This suggests that the VLM alone is insufficient to fully capture the user's intent and the semantic reasoning required for the task. 
Incorporating MLLMs leads to a substantial improvement in performance.
Among the evaluated models, ChatGPT-4.1 achieves the highest mAP across all evaluated ranks (top-K), demonstrating that larger, more advanced MLLMs are better at understanding a user's intent. 
Not surprisingly, the ``mini'' variants perform slightly worse than their full-sized counterparts, reflecting a trade-off between model capacity and retrieval accuracy. Overall, these results underscore the importance of strong semantic reasoning in the query fusion stage for training-free ZS-CIR.



\begin{table}[tb]
  \caption{Effect of MLLM choice on SQAF performance. Results are reported using CLIP ViT-G/14 as the VLM on the CIRCO test set. ``w/o MLLM'' indicates that the MLLM-generated description of the target image is not included.}
  \label{tab:ablation_mllms_circo_SQAF}
  \centering
  \begin{tabular*}{.6\columnwidth}{@{} @{\extracolsep{\fill} } l *{4}{c} @{} }
    \toprule
    MLLM                   & mAP@5          & mAP@10         & mAP@25         & mAP@50         \\
    \midrule
    w/o MLLM      & 18.05          & 19.02          & 20.59          & 21.42          \\
    ChatGPT-4.1            & \textbf{31.45} & \textbf{32.57} & \textbf{35.30} & \textbf{36.44} \\ 
    ChatGPT-4.1-mini       & 29.57          & 30.66          & 33.17          & 34.26          \\ 
    ChatGPT-4o             & 30.89          & 32.06          & 34.52          & 35.64          \\ 
    ChatGPT-4o-mini        & 27.01          & 27.61          & 30.18          & 31.22          \\ 
    \bottomrule
  \end{tabular*}
\end{table}

\paragraph{Effect of different MLLMs as rerankers}
As shown in Table~\ref{tab:ablation_mllms_circo_ebr}, we observe similar trends when MLLMs are used as rerankers in EBR as when they are employed in SQAF: MLLMs with stronger reasoning capabilities over visual and textual semantics, such as ChatGPT-4.1, consistently outperform the others.
Notably, ChatGPT-4.1-mini, despite its smaller size, achieves competitive performance, with an average mAP that is less than 1\% lower than ChatGPT-4.1.
This suggests that ChatGPT-4.1-mini offers a cost-effective alternative with minimal performance trade-off.
In contrast, ChatGPT-4o yields unsatisfactory results, 
likely due to its optimization for speed and efficiency, thereby compromising its visual-textual understanding and cross-image comparison capabilities.
Such a design choice therefore hinders ChatGPT-4o's effectiveness in reranking, as our EBR strategy relies on an MLLM capable of accurately interpreting user intent and performing cross-image reasoning to evaluate how well candidate images align with that intent.


\begin{table}[tb]
  \caption{Effect of different MLLMs used as rerankers. Results are reported using CLIP ViT-G/14 as the VLM and ChatGPT-4o within the SQAF module on the CIRCO test set.}
  \label{tab:ablation_mllms_circo_ebr}
  \centering
  \begin{tabular*}{.6\columnwidth}{@{} @{\extracolsep{\fill} } l *{4}{c} @{} }
    \toprule
    MLLM                & mAP@5          & mAP@10         & mAP@25         & mAP@50         \\
    \midrule
    w/o EBR & 30.89          & 32.06          & 34.52          & 35.64          \\
    ChatGPT-4.1         & \textbf{35.61} & \textbf{35.64} & \textbf{37.70} & \textbf{38.82} \\ 
    ChatGPT-4.1-mini    & 34.34          & 34.77          & 36.87          & 37.99          \\ 
    ChatGPT-4o          & 30.87          & 31.96          & 34.09          & 35.21          \\ 
    ChatGPT-4o-mini     & 24.71          & 26.50          & 29.13          & 30.25          \\ 
    \bottomrule
  \end{tabular*}
\end{table}

\begin{table}[tb]
  \caption{Effect of grid image size on MLLM-based reranking. Results are reported on the CIRCO test set using CLIP ViT-G/14 as the VLM, ChatGPT-4o within the SQAF module, and ChatGPT-4.1 as the reranker.}
  \label{tab:ablation_grid_reranking}
  \centering
  \begin{tabular*}{.6\columnwidth}{@{} @{\extracolsep{\fill} } l *{4}{c} @{} }
    \toprule
    Grid Image Size           & mAP@5          & mAP@10         & mAP@25         & mAP@50         \\
    \midrule
    w/o EBR & 30.89          & 32.06          & 34.52          & 35.64          \\
    $3 \times 3$        & \textbf{37.97} & \textbf{36.89} & \textbf{39.28} & \textbf{40.40} \\ 
    $4 \times 4$        & 35.61          & 35.64          & 37.70          & 38.82          \\ 
    $5 \times 5$        & 31.06          & 32.05          & 34.37          & 35.49          \\ 
    $6 \times 6$        & 24.31          & 24.90          & 27.49          & 29.00          \\ 
    \bottomrule
  \end{tabular*}
\end{table}

\paragraph{Effect of grid image size on MLLM-based reranking}
Table~\ref{tab:ablation_grid_reranking} shows how the grid size (\ie, the number of candidate images) affects the reranking performance. The 3×3 grid achieves the highest mAP scores, yielding an average improvement of 5.35\% over the one without reranking (\ie, w/o EBR).
The 4×4 grid also delivers a gain of 3.66\% mAP.
However, as the grid size increases to 5×5 and 6×6, performance degrades and even falls below that of w/o EBR. This decline is likely due to the increased visual complexity and the presence of more distractors in larger grids, which may overwhelm the MLLM’s reasoning capacity and hinder its ability to accurately assess image relevance to the query.

While the 3x3 grid achieves the best performance, we adopt the 4x4 grid as a practical choice. 
In real-world search scenarios, users are typically presented with a relatively large set of candidate images, and the 4×4 grid enables refinement over a broader pool of candidates (16 versus 9), thereby better reflecting practical usage. 
Furthermore, the ability to influence the top 16 ranks allows the reranker to have a greater effect on recall@K, which serves as the primary evaluation metric in most benchmarks.

\begin{figure}[tb]
    \centering
    \includegraphics[width=.7\textwidth]{figures/rerank_prompt_text_saqf.pdf}
    \caption{Example prompt for reranking with image captions generated in SQAF. The prompt guides the MLLM to reason over visual differences and update the ranking based on how well the candidates align with the user’s intent expressed as the captions.}
    \Description{}
    \label{fig:rerank_prompt_text_saqf}
\end{figure}

\begin{table}[tb]
  \caption{Comparison of different forms of user intent for reranking. Results are reported on the CIRCO validation set using CLIP ViT-G/14 as the VLM, ChatGPT-4o within the SQAF module, and ChatGPT-4.1 as the reranker.}
  \label{tab:ablation_rerank_inputs_results}
  \centering
  \begin{tabular*}{.8\columnwidth}{@{} @{\extracolsep{\fill} } l *{4}{c} @{} }
    \toprule
    User Intent                       & mAP@5          & mAP@10         & mAP@25         & mAP@50         \\
    \midrule
    Reference Image with Modification Text  & \textbf{35.05} & \textbf{35.62} & \textbf{37.37} & \textbf{38.26} \\
    Generated Target Captions               & 32.57          & 33.47          & 35.78          & 36.67          \\
    \bottomrule
  \end{tabular*}
\end{table}

\begin{figure}[tb]
    \centering
    \includegraphics[width=.6\columnwidth]{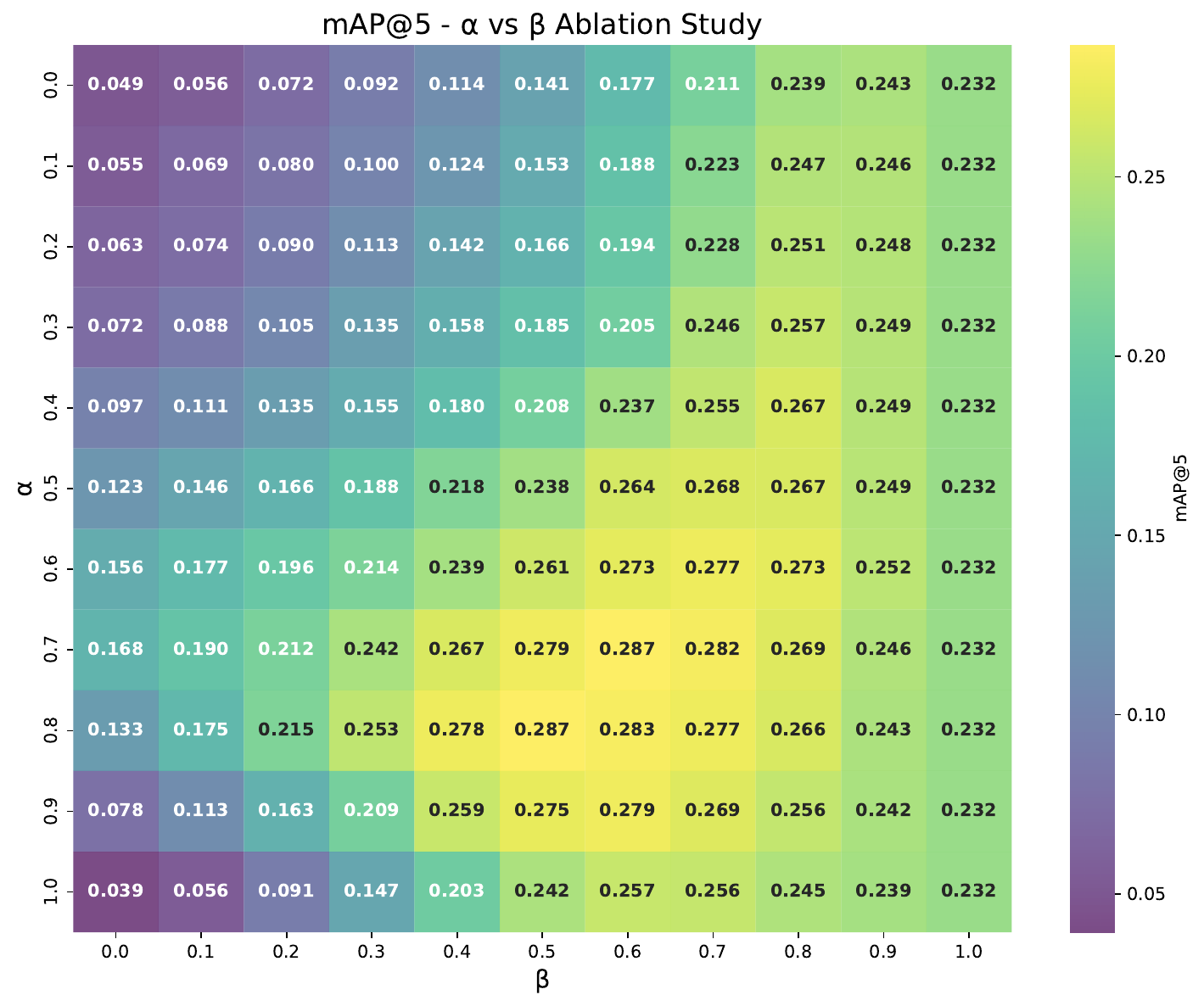}
    \caption{
Effect of the fusion hyperparameters $\alpha$ and $\beta$ on SQAF performance. $\alpha$ weights the relative contributions of the reference image and the modification text, and $\beta$ determines the influence of the MLLM-generated description in the final query. Results are reported using CLIP ViT-G/14 on the CIRCO validation set.
    }
    \Description{
Heatmap of mAP@5 scores on CIRCO validation set by varying $\alpha$ and $\beta$ values in Semantic Query-Augmented Fusion using CLIP ViT-G/14.
    }
    \label{fig:circo_heatmap}
\end{figure}

\paragraph{Comparison of different forms of user intent for reranking}
Our EBR ranks the candidate images based on user intent expressed through the reference image and modification text. As an alternative, we also consider representing user intent with the MLLM-generated captions from the SQAF module. To evaluate this, we use the prompt shown in Figure~\ref{fig:rerank_prompt_text_saqf} to guide the MLLM-based reranker with these captions and report results for different intent representations in Table~\ref{tab:ablation_rerank_inputs_results}.
As observed, using the reference image and modification text yields better results than relying solely on the MLLM-generated captions. This is likely because the original multimodal query offers precise, directly grounded guidance, whereas the captions, while semantically rich, may introduce irrelevant or misleading details.
In Section~\ref{sec:qualitative}, we further provide a qualitative analysis to better understand the effects.

\begin{figure}[tb]
    \centering
    \includegraphics[width=1\textwidth]{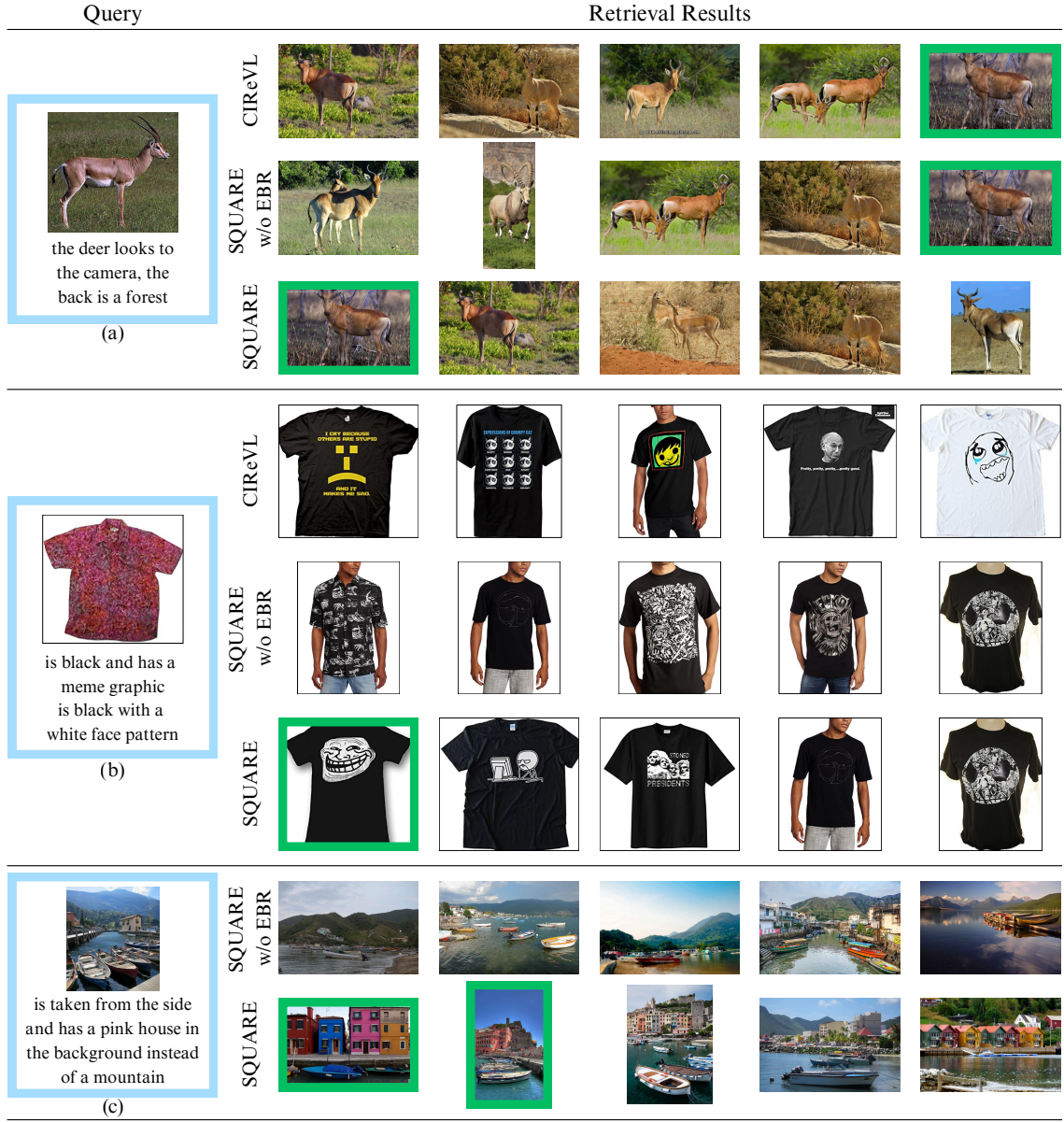}
    \caption{Qualitative comparisons of different retrieval methods on sample queries from (a) CIRR, (b) FashionIQ, and (c) CIRCO using CLIP G/14. Images highlighted with a green box denote the ground-truth target images.} 
    \Description{}
    \label{fig:qualitative_all_in_one}
\end{figure}

\paragraph{Effect of the fusion hyperparameters $\alpha$ and $\beta$ on SQAF performance} 
The heatmap in Figure~\ref{fig:circo_heatmap} shows how the final query representation, constructed using fusion weights $\alpha$ and $\beta$, influences global retrieval performance.
The parameter $\alpha$ in Equation~\eqref{eqn:init-fusion} controls the relative contributions of the reference image and the modification text, and $\beta$ governs the influence of the MLLM-generated description in the final composed query.

When the MLLM-generated description is excluded (\ie, $\beta = 0$), SQAF relies solely on the fusion of visual and textual embeddings from the VLM and produces inferior performance. This could be due to the fact that the VLM-fused representation lacks certain essential contextual descriptions.
As $\beta$ increases, the performance initially improves, suggesting that incorporating MLLM-generated descriptions enhances the understanding of the intended target image. However, beyond a certain point, further increasing $\beta$ leads to performance degradation. 
This indicates that the MLLM-generated captions still need to be constrained by the VLM-based query, as over-reliance on them may overshadow the VLM-based query intent or introduce noise that hinders accurate retrieval.
Interestingly, using the MLLM-generated description alone (\ie, $\beta = 1$) outperforms relying solely on the VLM-based embeddings. This observation implies that the MLLM is capable of producing high-level, semantically accurate descriptions that effectively capture the user’s intent.
Nonetheless, the best performance is achieved by combining MLLM-generated context with the VLM-based query, highlighting their complementary strengths.


\begin{figure}[tb]
    \centering
    \includegraphics[width=1\textwidth]{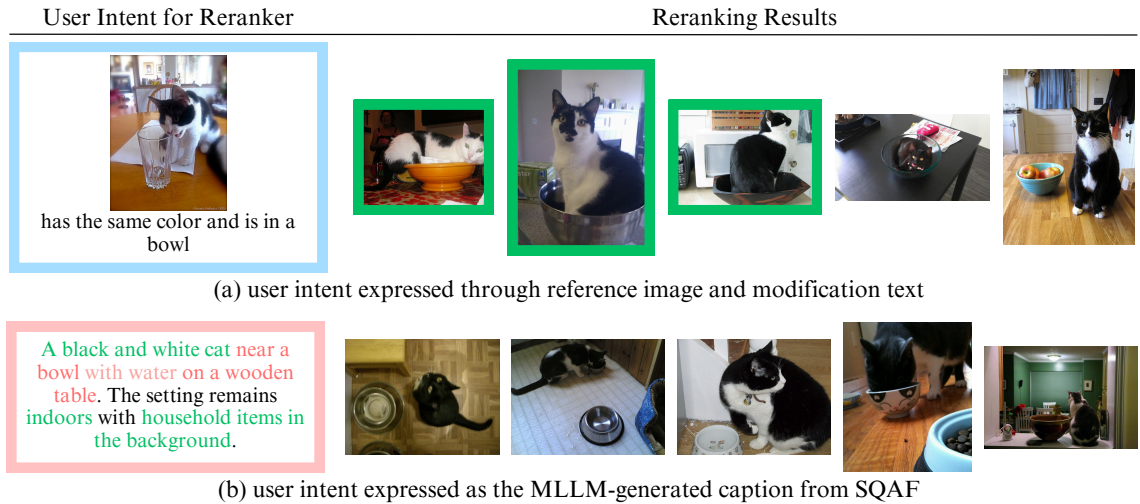}
    \caption{Qualitative comparison of different forms of user intent for reranking on CIRCO validation samples using CLIP G/14. (a) Results obtained using the reference image together with the modification text, which provide more reliable guidance for reranking. (b) Results relying solely on the MLLM-generated caption from SQAF. While MLLM captions provide high-level semantics, they may introduce errors (\eg, misinterpreting the cat as beside the bowl), leading to semantically incorrect reranking. Images highlighted with a green box denote the ground-truth target images.}
    \Description{}
    \label{fig:compare_rerank_context}
\end{figure}

\begin{figure}[tb]
    \centering
    \includegraphics[width=1\textwidth]{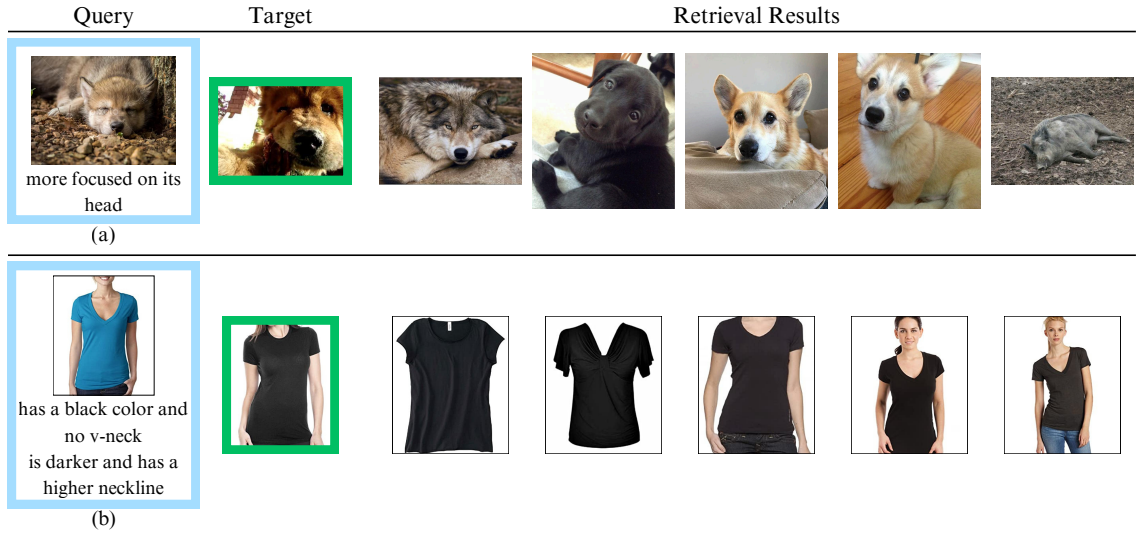}
    \caption{Failure cases of SQUARE on samples image using CLIP G/14. (a) Spatial reasoning limitation: the query from CIRR asks for the dog’s head to appear closer to the camera, but retrieved results mainly show frontal views rather than close-ups. (b) Compound attribute modification: the query from FashionIQ requests changing a blue V-neck shirt into a black non–V-neck shirt with a higher neckline. Our method fails to retrieve the annotated target image but still ranks semantically relevant alternatives among the top candidates. Images highlighted with a green box denote the ground-truth target images.}
    \Description{}
    \label{fig:failure_case}
\end{figure}

\subsection{Qualitative Analysis}
\label{sec:qualitative}

\paragraph{Visual comparison with other methods} 
As shown in Figure~\ref{fig:qualitative_all_in_one}, given a multimodal query, SQUARE retrieves the target image more effectively than other methods.
CIReVL often retrieves images that are semantically related to the query but fail to match the intended target.
Although SQUARE w/o EBR is able to retrieve the target, the target often appears at a lower rank, especially in challenging scenarios with highly similar gallery images (\eg, CIRCO), where SQAF alone struggles to capture fine-grained distinctions.
In contrast, SQUARE demonstrates that incorporating the EBR module, which leverages the enhanced multimodal reasoning ability of the MLLM, consistently promotes the target image to a higher rank, clearly highlighting the benefit of EBR for improved retrieval performance.

\paragraph{Qualitative comparison of different forms of user untent for reranking}
In Section~\ref{sec:ablation}, we quantitatively compared different contexts used for reranking and found that using the reference image together with the modification text outperforms relying solely on the \textit{MLLM-generated captions} from SQAF. Here, we provide a qualitative analysis.
As shown in Figure~\ref{fig:compare_rerank_context}, although MLLM-generated captions can offer fine-grained semantic cues to guide reranking, they may also introduce inaccurate or misleading descriptions. For instance, the generated caption misrepresents the relationship between the `cat' and the `bowl',
describing the cat as being \textit{beside} the bowl rather than \textit{in} it.
This conflict with the modification text causes the retrieval model to favor images where the cat is next to the bowl, instead of correctly retrieving images where the cat is inside the bowl.


\paragraph{Failure cases} 
Despite achieving promising results, our method still exhibits certain limitations in specific cases. As shown in Figure~\ref{fig:failure_case}(a), the modification requests a composition focused on the dog's head, with the target image depicting the dog's head positioned closer to the camera. 
Since most of the retrieved images show the dog's face simply facing the camera rather than in a close-up, this failure suggests that the model may be limited in its spatial reasoning. We also conjecture that the modification text is somewhat ambiguous, which, combined with the model's limitations, prevents it from fully grasping the requested change in relative distance and perspective.

Figure~\ref{fig:failure_case}(b) illustrates a scenario in which the user modifies multiple attributes of a fashion image simultaneously. For example, the query specifies the change of a blue V-neck shirt into a black non–V-neck shirt with a higher neckline. Because such compound modifications require capturing subtle interactions between color, style, and neckline, our method fails to retrieve the annotated target image. 
However, it still ranks other alternative images that satisfy the semantic request as the top candidates.
Thus, even when the exact target is not retrieved, our method returns semantically relevant results in challenging scenarios.

\section{Conclusion}
In this paper, we presented SQUARE, a novel two-stage training-free framework for ZS-CIR. In the first stage, Semantic Query-Augmented Fusion (SQAF) enhances the composed query by integrating an MLLM-generated semantic caption with the VLM-based embedding query, thereby improving global retrieval quality. In the second stage, Efficient Batch Reranking (EBR) exploits the reasoning capability of MLLMs to jointly compare all candidate images in a single forward pass, leading to more accurate and semantically coherent ranking.
Extensive experiments demonstrate that SQUARE achieves state-of-the-art or near state-of-the-art performance across multiple benchmarks. Notably, our method yields substantial improvements when applied with smaller VLMs (\eg, CLIP B/32), in some cases even surpassing results that were previously achievable only with larger models (\eg, CLIP L/14).
Our experiments also highlight the strong potential of recent MLLMs, such as ChatGPT-4.1, as highly effective rerankers, with even lightweight variants delivering competitive performance.
We hope that SQUARE, as a simple yet effective framework, not only advances research in training-free ZS-CIR but also inspires further exploration in this direction.

\bibliographystyle{ACM-Reference-Format}
\bibliography{refs}

\end{document}